\documentclass[letterpaper]{article}
\usepackage[margin=1in]{geometry}
\usepackage{amsmath} 
\usepackage{amssymb}  
\usepackage{graphicx}
\usepackage{subcaption}
\usepackage{bm}
\usepackage[nospace]{cite}
\usepackage[font=footnotesize]{caption}
\usepackage[colorinlistoftodos]{todonotes}
\usepackage{comment}
\usepackage[colorinlistoftodos]{todonotes}

\title{
Vision-Based Autonomous Vehicle Control\\ using the Two-Point Visual Driver Control Model
}

\author{\thanks{This work was supported by CPS-1544814 award from NSF and a PURA award from Georgia Tech.
		K.Okamoto was also partially supported by Funai Foundation for Information Technology and Ito Foundation USA-FUTI Scholarship.}
		Justin Zheng\thanks{J. Zheng is with Waymo. This work was done while at the School of Electrical and Computer Engineering,
        Georgia Institute of Technology, GA 30332-0150, USA. Email:~justin\_zheng@gatech.edu}\qquad Kazuhide Okamoto\thanks{K. Okamoto is with Zoox. 
        This work was done while at the School of Aerospace Engineering, Georgia Institute of Technology, Atlanta, GA 30332-0150, USA. Email:~kazuhide@gatech.edu}\qquad Panagiotis Tsiotras\thanks{P. Tsiotras is with the School of Aerospace Engineering and the Institute for Robotics \& Intelligent Machines, Georgia Institute of Technology, Atlanta, GA 30332-0150, USA. Email: tsiotras@gatech.edu}}

\begin{document}

\maketitle
\thispagestyle{empty}
\pagestyle{empty}

\begin{abstract}
This work proposes a new self-driving framework that uses a human driver control model, whose feature-input values are extracted from images using deep convolutional neural networks (CNNs).
The development of image processing techniques using CNNs along with accelerated computing hardware has recently enabled real-time detection of these feature-input values. 
The use of human driver models can lead to more ``natural'' driving behavior of self-driving vehicles.
Specifically, we use the well-known two-point visual driver control model as the controller, and we use a top-down lane cost map CNN and the YOLOv2 CNN to extract feature-input values.
This framework relies exclusively on inputs from low-cost sensors like a monocular camera and wheel speed sensors.
We experimentally validate the proposed framework on an outdoor track using a 1/5th-scale autonomous vehicle platform. 
\end{abstract}

\section{INTRODUCTION}

\begin{figure}[htb]
	\centering
	\includegraphics[width=0.7\columnwidth]{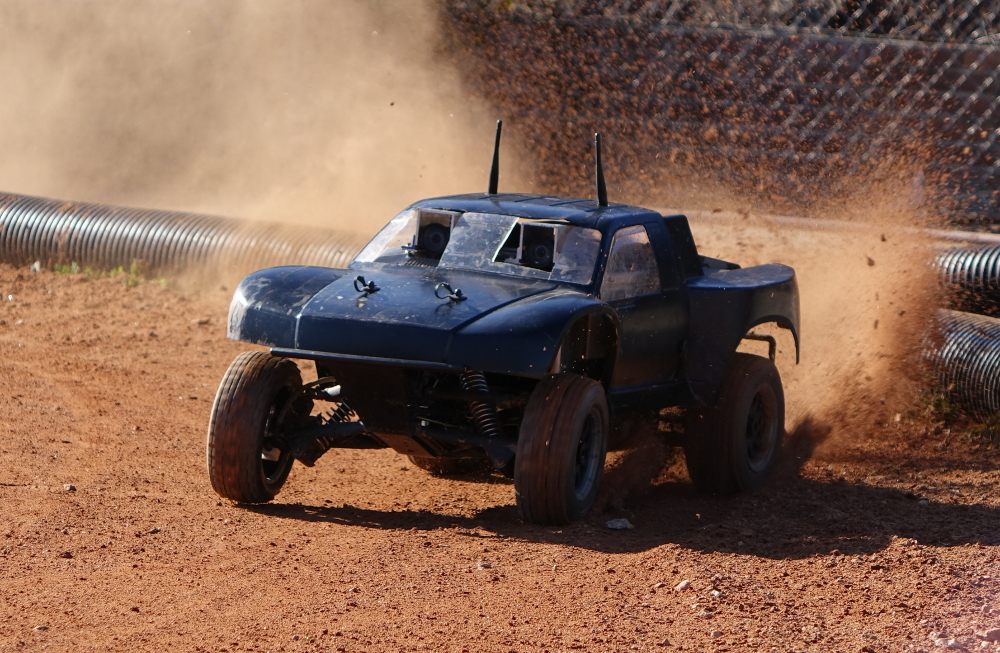}
    \caption{\small{AutoRally, a 1/5th scale autonomous vehicle platform.}}
    \label{fig:autorally}
\end{figure}

Since the DARPA Grand Challenge competition \cite{thrun2006stanley}, autonomous driving has been actively researched. 
However, many of the previously proposed control frameworks are not designed to capture natural tendencies of human drivers, implying that their behavior can be significantly different from human driving.
One goal of self-driving is to maintain control similarly to humans so that the vehicle's motion is interpretable and comfortable.
While mathematically formulating the similarity between an autonomous driver and a human driver can be difficult, we conjecture that human driver control models can help achieve similarity to normal human driver behavior.
The aim of this paper is to demonstrate the use of a human-driver-control model within the control loop of a physical autonomous vehicle platform, with a focus on maintaining lateral stability.
Specifically, we use the two-point visual driver control model (TPVDCM) described in~\cite{salvucci2004two} to control a 
vehicle, and we implement this controller on the AutoRally platform \cite{autorally}, a 1/5th-scale autonomous vehicle for research in advanced perception and control (Fig.~\ref{fig:autorally}).
We show that a convolutional neural network (CNN) such as a top-down network for predicting the lane cost map \cite{drews2017deepdriving,drews2018cnn} or YOLOv2 \cite{redmon2017yolo} can be used to extract visual cues, or feature-input values, for the TPVDCM.

The contributions of this paper are two fold. 
First, we propose a methodology for using existing convolutional neural networks to extract feature-input values for the TPVDCM in real-time, thus enabling control of a physical vehicle using only low-cost sensors like a monocular camera and wheel speed sensors.
Second, we provide experimental evidence that the TPVDCM can provide lateral control using the AutoRally platform driving on an outdoor track.
To the best of our knowledge, this is the first work that uses a human driver control model to steer a real vehicle.

\section{RELATED WORK}

\begin{figure}[htb]
	\includegraphics[width=\columnwidth]{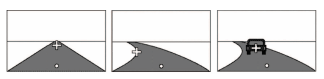}
	\hspace*{34pt} (a) \hspace{62pt} (b) \hspace{62pt} (c)
    \caption{\small{The two-point visual driver control model uses a near feature point (white dot) and a far feature point (white cross) to establish lateral stability and lane tracking; from \cite{salvucci2004two}.}}
    \label{fig:twoPoint}
\end{figure}

The objective of this paper centers around capturing the tendencies of human drivers in order to maintain lateral stability using purely a vision-based approach without the need for localization or mapping.
Human drivers are known to use near and far indicator points in their field of view to assist with driving tasks such as lane following, lane switching, and position correction~\cite{salvucci2004two}. 
A simulation study~\cite{donges1978two} showed that steering could be described as a two-level control problem, in which nearby points on the road are used for identifying the immediate error in the vehicle position, while far away points are used to anticipate control actions for an upcoming segment of the road (Fig.~\ref{fig:twoPoint}). 
The TPVDCM described in \cite{salvucci2004two} uses these points as an input to a closed-loop feedback control system, which then outputs a control value to the steering wheel.
This model was supported by behavioral studies \cite{mars2008behavorial}.
In \cite{sentouh2009sensorimotor}, the TPVDCM was extended to include a two-level visual strategy and high-frequency compensation based on kinesthetic feedback to better model human driver characteristics.
Reference \cite{zafeiropoulos2014twopoint} uses the TPVDCM as part of a driver steering assist system, and incorporates steering column dynamics to control a simulated vehicle.
In \cite{YouLuTsi:hms16}, the authors show that the TPVDCM can be used to design better advanced driver assistance systems (ADAS).
A more sophisticated anticipatory channel for the TPVDCM based on model predictive control (MPC) is described in \cite{okamoto2016}.

While the TPVDCM has been utilized to investigate and understand human driver actions, to the authors' knowledge, it has not been implemented on real vehicle platforms. 

In order to implement the TPVDCM on the AutoRally platform, feature-input values to the control model must be extracted in real-time.
One approach that can be used to detect both the near and far indicator points is lane centerline detection.
Lane detection has been a well-studied problem over the last several years.
Reference \cite{huang2009lanedetect} combines an artificial neural network (ANN), particle filtering, and RANSAC to detect lane markings.
More recent works \cite{li2017lanedetectdnnrnn,wang2018cnn} show impressive results for using CNNs to detect lane markings in real-time.
Reference \cite{drews2017deepdriving} describes a fully convolutional network designed to predict a lane cost map of the area directly in front of a vehicle using a single RGB image from a monocular camera as an input, where the minimum cost is at the lane centerline.
In \cite{drews2017deepdriving}, it was shown that the use of a CNN for directly predicting a top-down cost map achieves better results than predicting a cost map in the driver-point-of-view image.
Reference \cite{drews2018cnn} improves upon the CNN used in \cite{drews2017deepdriving} by introducing an encoder-decoder architecture and LSTM architecture to generate the lane cost map.
A lane cost map can be used to identify the lane centerline, and two points along the lane centerline can be used as near and far indicator points for the TPVDCM (Fig.~\ref{fig:twoPoint}(a)). A lane cost map would also show the inner boundary along a turn, which can be used as the far point for the TPVDCM (Fig.~\ref{fig:twoPoint}(b)).

An object detection CNN can be used to detect a leading vehicle in the camera image, which can subsequently be used as the far indicator point (Fig.~\ref{fig:twoPoint}(c)). 
YOLOv2 \cite{redmon2017yolo} shows significant improvement in computation time for object detection tasks compared to other methods such as Fastest DPM \cite{yan2014dpm} and Fast R-CNN \cite{girshick2015rcnn}, which do not meet the criteria for real-time detection.
Inspired by a human's ability to view the world and rapidly estimate the location and relationships of objects in sight, YOLOv2 provides a robust pipeline to detect objects at a rate compatible with the TPVDCM. 
The combination of a top-down CNN for lane cost map prediction and YOLOv2 with the TPVDCM establishes the vision-based feature extraction used in this paper.

\section{SYSTEM MODEL AND CONTROLLER}

\begin{figure*}[htbp] 
	\centering
	\includegraphics[width=0.75\textwidth]{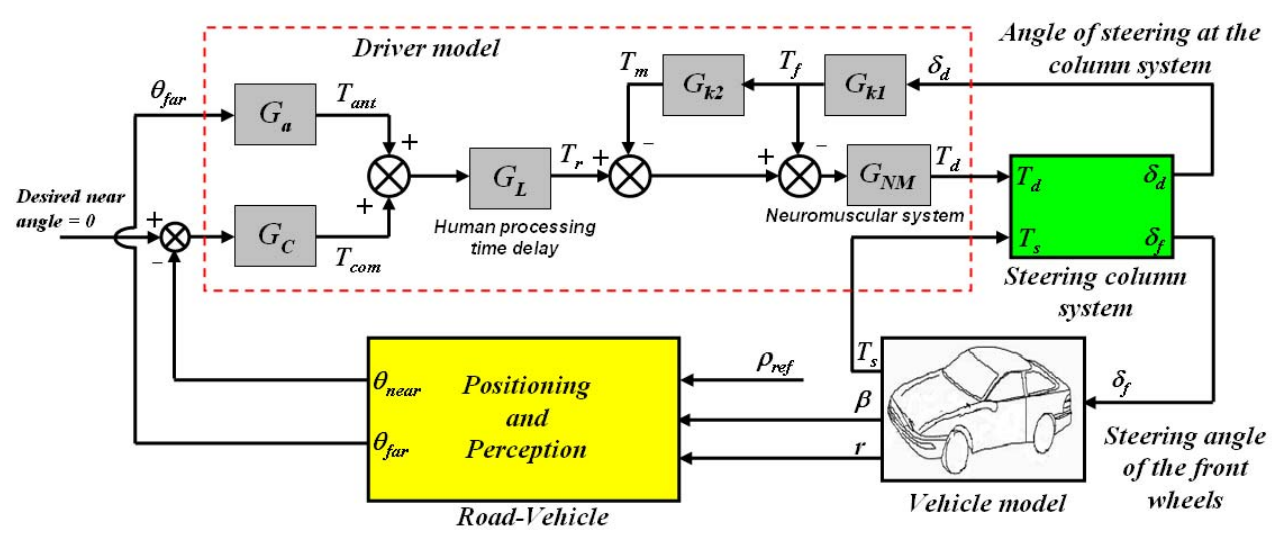}
    \caption{\small{Human-vehicle-road closed-loop system; from \cite{sentouh2009sensorimotor}. The visual feature-input point extraction in this paper composes the "Positioning and Perception" block (yellow). $\theta_{\rm{near}}$ and $\theta_{\rm{far}}$ are the angles between the vehicle's heading and these feature-input points, which are inputs to the driver model (dotted red). The output of the driver model is a torque that is applied to the vehicle.}}
    \label{fig:driverModel}
\end{figure*}

This section describes the TPVDCM used in this work, which is based on the sensorimotor TPVDCM \cite{sentouh2009sensorimotor}, shown in Fig.~\ref{fig:driverModel}; see also Fig.~\ref{fig:theta_Near}.
The angles $\theta_{\rm{far}}$, $\theta_{\rm{near}}$, and $\delta_d$ are the inputs to the driver model, where $\theta_{\rm{far}}$ is the angle between the vehicle's heading and the far” point of interest (e.g., tangent point on curved roads, vanishing point on straight roads, vehicle in front in traffic, etc), and $\theta_{\rm{near}}$ is the angle between the vehicle's heading and the “near” point of interest. 
The angle $\delta_d$ is the steering column angle. The transfer function $G_a(s)$ corresponds to the anticipatory control of a human driver acting on $\theta_{\rm{far}}$. 
The transfer function $G_c(s)$ corresponds to the compensatory control of a driver acting on $\theta_{\rm{near}}$. The torques $T_{\rm{com}}$ and $T_{\rm{ant}}$ are the steering torques associated with compensatory and anticipatory control, respectively. The transfer function $G_L(s)$ corresponds to the processing delay of the driver. 
The transfer functions $G_{k1}(s)$ and $G_{k2}(s)$ correspond to the driver's kinesthetic perception of the steering angle $\delta_d$. The transfer function $G_{\rm nm}(s)$ corresponds to the neuromuscular response delay of the driver's arms. 
The output torque $T_{\rm{dr}}$ is the torque applied on the steering wheel. 
In practice, kinesthetic perception feedback is removed from the model since it has no significant effect on model performance \cite{zafeiropoulos2014twopoint}.

\section{Convolutional Neural Networks for Feature-Input Extraction}
\subsection{Top-Down Lane Prediction}

\subsubsection{Architecture}

\begin{figure}[htb]
    \centering
	\includegraphics[width=\columnwidth]{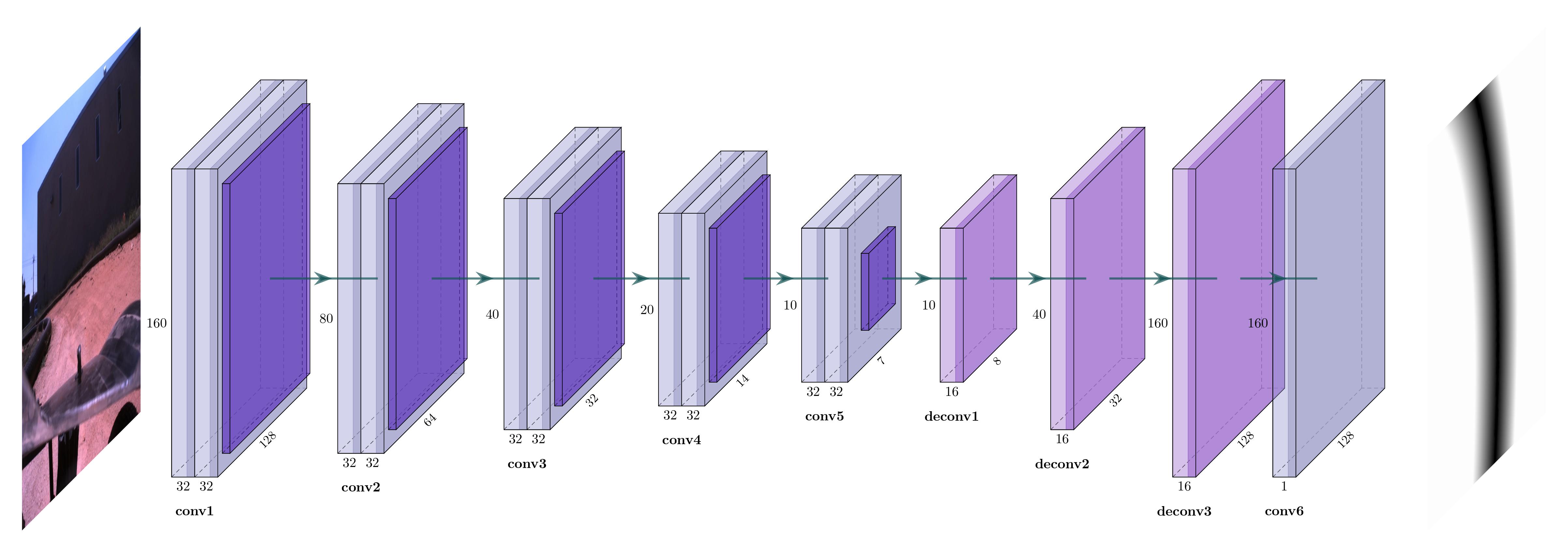}
    \caption{\small{Architecture for predicting a top-down lane cost map, with an input image and output cost map of size 160 x 128. Convolutional layers and average pooling layers compose the encoder portion of the network, and deconvolution layers upsample the feature maps to a grayscale representation of the lane cost map. This network is based on the encoder-decoder architecture from \cite{drews2018cnn}. }}
    \label{fig:deepdrivingnet}
\end{figure}

We use a modified version of the encoder-decoder network architecture from \cite{drews2018cnn} without the fully-connected bottleneck layers to output a dense top-down lane cost map (Fig.~\ref{fig:deepdrivingnet}). The input to the network is a 160 x 128 RGB image, and the output is a 160 x 128 grayscale cost map image.
Since the output cost map image shows the lane centerline with the lowest cost, we can use points along the lane centerline as feature-input values to the TPVDCM.

\subsubsection{Automatic Labeling}

Similarly to \cite{drews2017deepdriving} and \cite{drews2018cnn}, labels are automatically generated by driving the AutoRally platform manually around the track while recording camera images and running a localization algorithm \cite{dellaert2012gtsam} that makes use of GPS and IMU to get centimeter-level pose accuracy.
A top-down ground truth image of the lane can be generated for each estimated pose 
(Fig.~\ref{fig:deepdrivingoutput}).
The values of pixels in a label is lowest along the lane centerline, and scale quadratically as pixels get farther from the lane centerline. 
This labeling scheme is used as opposed to a binary classification of pixels, since a cost map reveals the lane centerline at the lowest cost, while also showing the lane boundaries. 
The top-down cost map output is generated assuming the ego vehicle is located at the bottom-center of the image.

\subsubsection{Training}
Data was aggregated from 15 test runs spanning two different days and 30 minutes of driving, resulting in approximately 80,000 examples. The data was split into approximately 70,000 examples for training and 10,000 examples for testing.

Our training process follows roughly the same procedure as in \cite{drews2018cnn}.
The network uses the Adam optimizer \cite{kingma2014adam} and is trained using the Tensorflow \cite{tensorflow2015-whitepaper} framework with an L1 pixel-wise loss function and a mini-batch size of 10 images.
The pixel-wise loss applies to the entire image, as opposed to \cite{drews2017deepdriving}, where loss is only computed for pixels on and near the track location. This is done to reduce noisy artifacts that appear outside the track boundaries if loss is not computed across the entire image.
For a training image, each color channel is multiplied by a normally distributed random variable between 0.8 and 1.05. Images were captured using a PointGrey Flea3 color forward-facing camera on the AutoRally platform at 1280 x 1024 resolution, and downsampled to 160 x 128. The network was trained for approximately 100,000 iterations. 

\subsubsection{Network Performance}
\begin{figure}[htb]
	\includegraphics[width=\columnwidth/2]{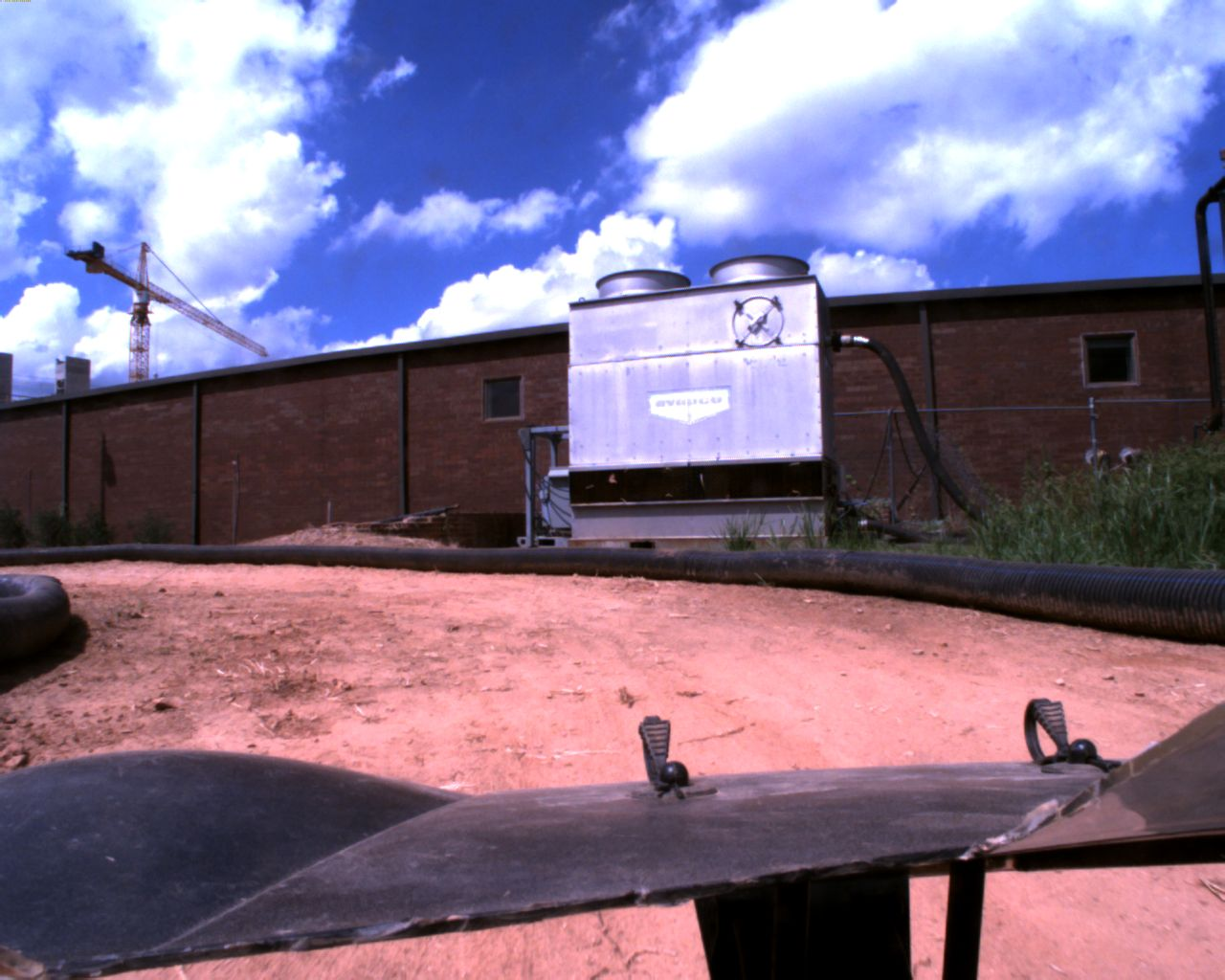}\includegraphics[width=\columnwidth/2]{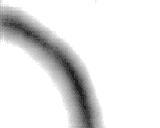}
    \caption{\small{Example input image [left] and corresponding top-down lane cost map prediction [right].}}
    \label{fig:deepdrivingoutput}
\end{figure}

This network runs at 40 Hz (at 40 Hz camera frame rate) on the AutoRally platform, which has an Nvidia GTX 1050Ti GPU. To evaluate the performance of the network, we compute a score as $score = 1 - error$, where the error is the L1 loss.
The network achieved an average score of 0.97 across the test set, which is higher than \cite{drews2017deepdriving} because error is computed across all pixels, as opposed to only the pixels that are on the track.
An example output is shown in Figure \ref{fig:deepdrivingoutput}.

\subsection{Vehicle Detection} \label{vehicle_detection}
\subsubsection{Architecture}
For the TPVDCM, we wish to use object detection to identify a ``far” point to use as a reference for anticipatory control. 
A convenient point to use is a vehicle driving along the road in front of the ego vehicle. This is equivalent to a human driver using the position of a leading vehicle to anticipate the road curvature ahead. 
We can use vehicle detection to identify a car in front of the AutoRally platform and use the heading angle to this car as $\theta_{\rm{far}}$. 

We use the YOLOv2 network architecture from \cite{redmon2017yolo} and ROS package \cite{bjelonicYolo2018} for object detection because it can predict bounding boxes and class probabilities of objects in real-time.
The input to the YOLOv2 network is a $416 \times 416$ RGB image.
YOLOv2 uses the Darknet framework \cite{redmond2013darknet} for implementation.
Reference \cite{redmon2017yolo} provides a trained model for the PASCAL Visual Object Classes (VOC) detection task \cite{everingham2015pascal}, which contains a ``car" class.
However, the performance of this model suffers in our case since the appearance of the AutoRally vehicle is not in the distribution of vehicles present in the VOC dataset.
The AutoRally platform is a 1/5th-scale vehicle, and is also driving on an off-road surface.
We therefore used the model from \cite{redmon2017yolo} as a benchmark, and continued training this model using labeled images of the AutoRally platform to improve detection performance.

\subsubsection{Automatic Labeling}
\begin{figure}[th]
\centering
\begin{subfigure}{.5\columnwidth}
\centering
\includegraphics[width=\columnwidth]{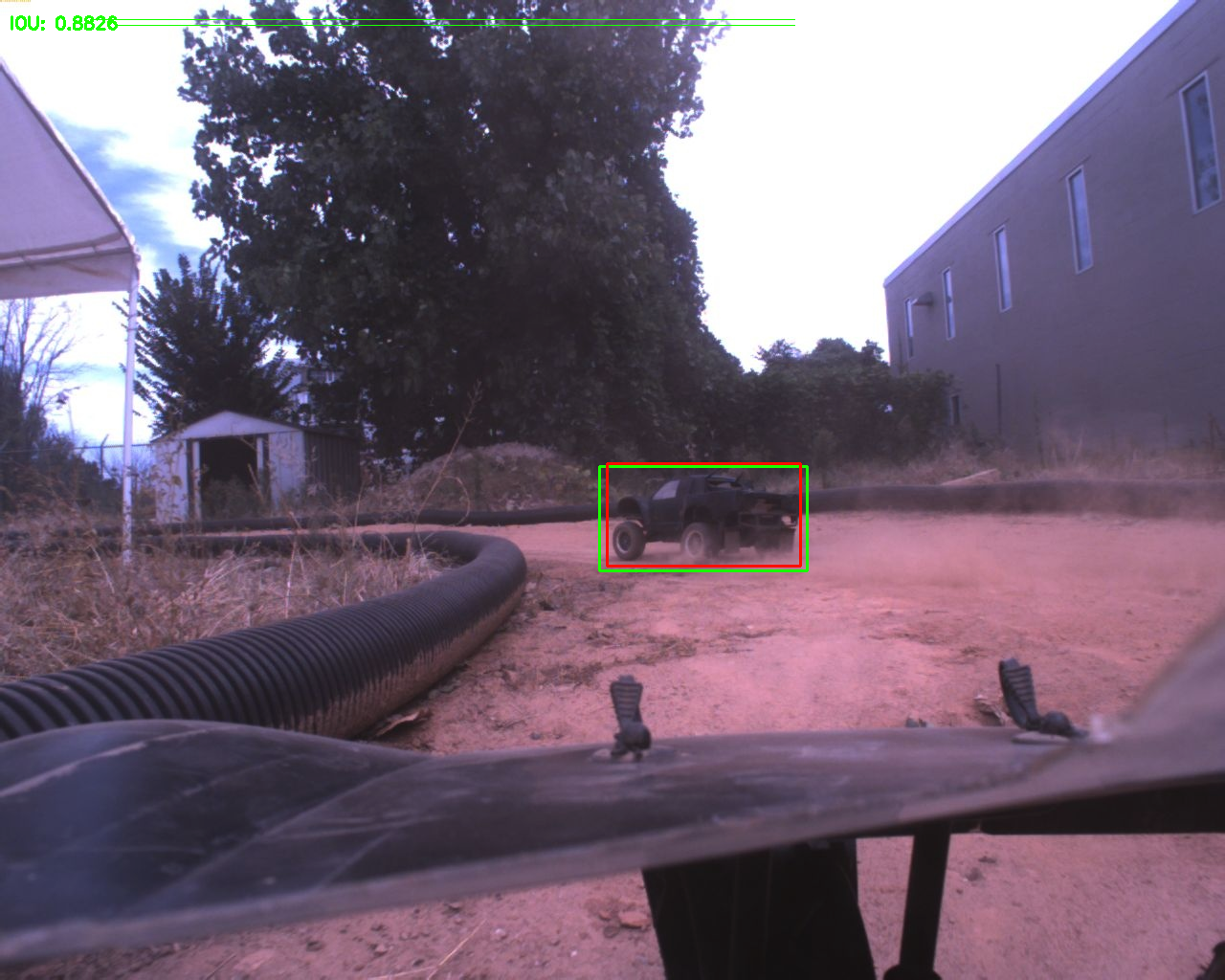}
\end{subfigure}%
\begin{subfigure}{.5\columnwidth}
\centering
\includegraphics[width=\columnwidth]{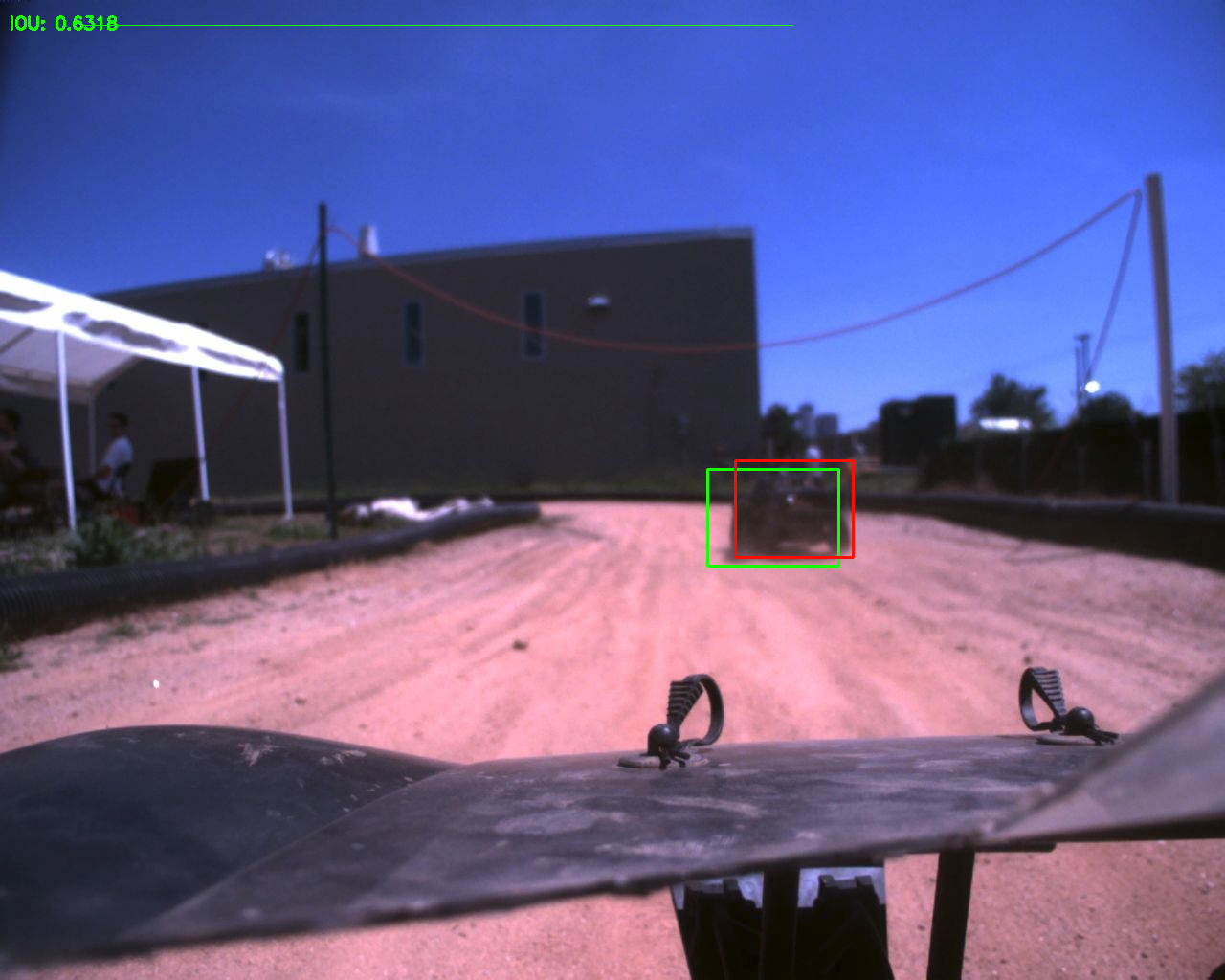}
\end{subfigure}\\[1ex]
\caption{\small{The ground truth bounding box generated with the automatic labeling tool (green), and the detected bounding box (red).}}
\label{fig:detection}
\end{figure}

Similarly to other deep neural networks, in order to successfully detect objects, YOLOv2 requires a significant number of training images.
Since we are tracking another AutoRally vehicle as the ``far" point for the TPVDCM, we only need images with a labeled AutoRally vehicle for training the YOLOv2 network. 
Training images were collected using an automatic labeling tool with no human supervision.
This approach requires one AutoRally platform to be driving in front of a second, trailing vehicle, with the trailing vehicle recording images. 
When the leading vehicle is in the field of view of the trailing vehicle's camera, the automatic labeling tool uses the pose estimates of both vehicles to map the position of the leading vehicle onto the image recorded by the trailing vehicle's camera at that moment (Fig.~\ref{fig:detection}). 
The following equation was used to identify the pixel representing the center of a vehicle in an image

\begin{equation}
\bm{r} \sim K G_{\rm{Trailing}}^{\rm{Cam}}G_{\rm{World}}^{\rm{Trailing}}q^{\rm{World}},
\end{equation}
where $\bm{r} = (x,y)$ is the pixel value, $K$ is the camera matrix containing the intrinsic parameters of the trailing vehicle's camera, $G_{\rm{Trailing}}^{\rm{Cam}}$ is the projection matrix that takes a point in the trailing vehicle's pose reference frame and projects the point to the vehicle's camera reference frame, $G_{\rm{World}}^{\rm{Trailing}}$ is the projection matrix that takes a point in the world reference frame and projects it to the trailing vehicle's pose reference frame.
$q^{\rm{World}}$ is a homogeneous point that represents the position of the leading vehicle. 
Since we know the dimensions of the AutoRally platform, we can also use Eq.~(3) to map the world coordinates of the leading vehicle's eight 3D bounding box corners onto an image, and subsequently draw a bounding box around the outermost points.

In order to obtain the matrices $K$ and $G^{\rm{Cam}}_{\rm{Trailing}}$, we used the Kalibr calibration tool \cite{maye2013selfsupervisedCF}. 
AutoRally uses the factor graph localization toolbox GTSAM \cite{dellaert2012gtsam} to estimate a vehicle's pose and velocity based on GPS, IMU, and wheel-speed signals.
This pose estimate is used to obtain the transformation from the world reference frame to the leading vehicle's pose reference frame, $G_{\rm{World}}^{\rm{Trailing}}$. 
GTSAM is also used to estimate the position of the leading vehicle in the world frame, $q^{\rm{World}}$.

\subsubsection{Training}

We initialize the network using the trained weights from \cite{redmon2017yolo} for the VOC detection task. 
After modifying the network to predict a single class instead of the 20 classes of VOC, we continued training the entire network on images of the AutoRally vehicle collected and labeled using the automatic labeling process. 
Running the automatic labeling tool across several test run videos produced 7,000 training images with the AutoRally platform, and 1,725 testing images.
A relatively low number of additional training images is required, since the network already has encoded some information about vehicle features.
The loss function used during training is a function of the error in the predicted bounding box center coordinate and size, which is explained further in \cite{redmon2017yolo}. 
Data augmentation methods like random cropping and color shifts are also used to avoid overfitting.
To train the network for detection of the AutoRally vehicle, a batch size of 4, momentum of 0.8, learning rate of $1.0\times10^{-4}$, and weight decay of $5.0\times10^{-4}$ are used for 10,000 iterations.

\subsubsection{Network Performance}

This network runs at 13 Hz on the AutoRally platform.
To assess the performance of the YOLOv2 network trained on VOC versus the network trained with additional images of the AutoRally platform, we compare three metrics: average intersection over union (IOU), recall, and precision.
We use a confidence threshold of 0.15 for our evaluation and during experiments.

The YOLOv2 network achieves an average IOU of 0.41 across the test dataset, whereas our network achieves an average IOU of 0.50.
Note that this IOU could be further improved with higher quality labels, since using the automatic labeling method described earlier can be prone to errors due to miscalibration and incorrect pose estimates.

To compute recall and precision, true positives are defined as detected bounding boxes with an IOU over a threshold of 0.15, and false positives are defined as detected bounding boxes that do not overlap with the ground truth bounding box above the IOU threshold.
Our network achieved a recall of 0.97 and precision of 0.92, whereas the YOLOv2 network trained on just VOC achieves a recall of only 0.30 with the same precision.
We can use a low confidence threshold and IOU threshold for our experiments because it is important to achieve a high recall of vehicle detections so that the vehicle being controlled with the TPVDCM can reliably follow a leading vehicle.
We also assumed that there is at most one vehicle in the image so that it is unlikely our network predicts false positives.
Fig.~\ref{fig:detection} contains images comparing the ground truth and the bounding box predicted by YOLOv2 on the testing set.

\section{Two-Point Visual Controller on the AutoRally Platform}

\subsection{Ground Truth Feature Extraction} \label{baseline}

In order to assess the two-point visual controller's performance on the AutoRally vehicle, we first describe the process to determine $\theta_{\rm{near}}$ and $\theta_{\rm{far}}$ using a hard-coded ground truth centerline of the track instead of measuring them using vision-based methods.
This process requires localization of the AutoRally platform on the track, which is accomplished by using GTSAM and the GPS, IMU, and wheel speed sensors onboard the AutoRally platform.
The purpose of this process is to establish a baseline for controlling the AutoRally platform using the TPVDCM assuming near-perfect feature-input values, and compare this performance against our vision-based methods.

The track used for testing is roughly ellipsoidal in shape, with a uniform three meter wide driving surface and a maximum outer diameter of 30 meters.
In order to define the true “center” of the track, we drove the vehicle one lap along the track centerline while recording its position estimate, which is obtained using GTSAM.
The resulting position coordinates can be saved as a point cloud.
This centerline can then be used to extract $\theta_{\rm{near}}$ and $\theta_{\rm{far}}$.

\begin{figure}[th]
	\centering 
	\includegraphics[width=0.7\columnwidth]{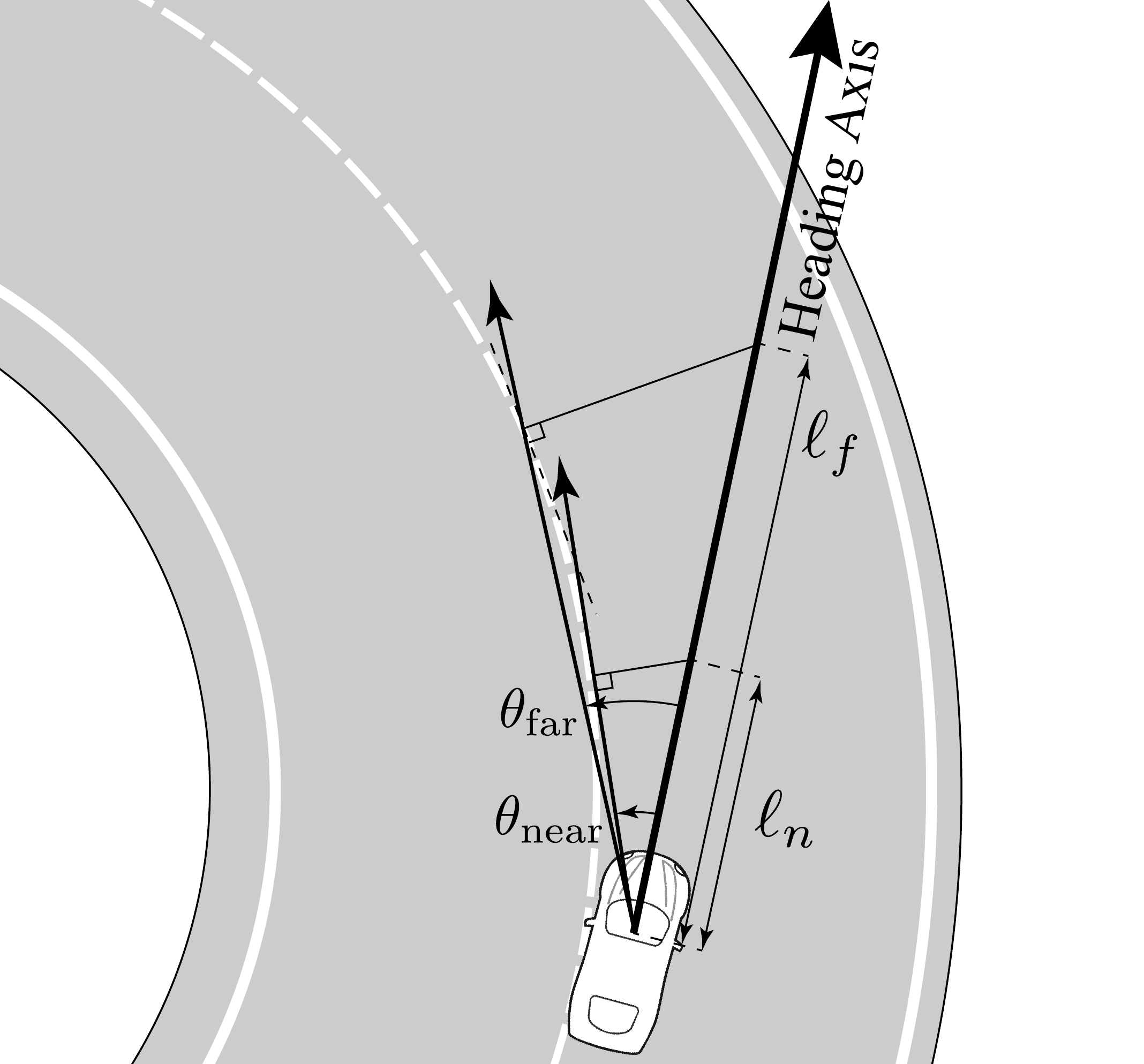}
    \caption{\small{Calculation of $\theta_{\rm{near}}$ and $\theta_{\rm{far}}$ using ground truth feature extraction.}}
    \label{fig:theta_Near}
\end{figure}

We approximate $\theta_{\rm{near}}$ as follows (see Fig.~\ref{fig:theta_Near}):
a) Obtain a pose estimate of the AutoRally platform;
b) Identify a point that is $\ell_n = $ 1 m away along the heading axis of the vehicle;
c) Find the nearest neighbor of this point along the track centerline;
d) Compute $\theta_{\rm{near}}$ from
\begin{equation} \theta_{\rm{near}}\approx\arctan{\left(\frac{y_{nn} - y_{\tt{car}}}{x_{nn} - {x_{\tt{car}}}}\right)},
\end{equation}
where $nn$ symbolizes the coordinate of the nearest neighbor in the centerline, and $\mathtt{car}$ represents the coordinate of the point along the car's heading axis.

In order to approximate $\theta_{\rm{far}}$ for lane-following tasks, we follow the same process as Step 2 with a point farther away, where $\ell_f = $ 3 m away.

\subsection{Vision-Based Feature Extraction}

The previously described CNNs for top-down lane prediction and vehicle detection are used to extract feature-input values for the TPVDCM.
We describe how the CNNs are used to extract feature-input values for the TPVDCM.

\subsubsection{Top-Down Lane Prediction} \label{top_down_setup}
The top-down lane prediction network can be used to identify $\theta_{\rm{near}}$ and $\theta_{\rm{far}}$.
The output of this network is a top-down view of the lane, where the vehicle's position is at the bottom-center of the image.
The dimensions of the network output are 160 x 128, where 15 pixels represent 1 meter.

\begin{figure}[th]
	\centering 
	\includegraphics[width=0.5\columnwidth]{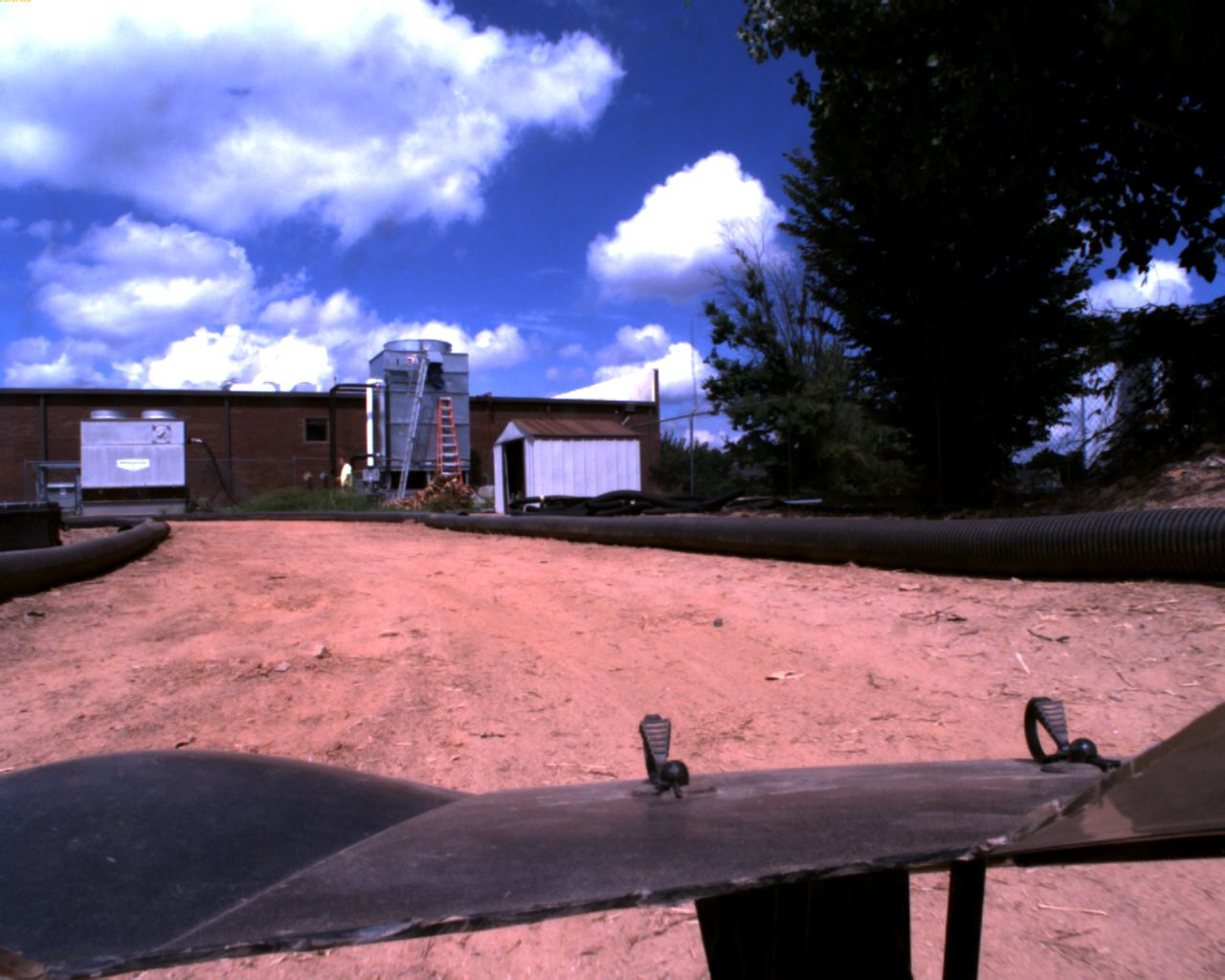}\includegraphics[width=0.5\columnwidth]{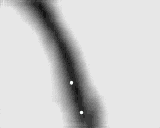}
    \caption{\small{Input image to top-down lane cost map prediction network [left] and output of network [right] with white dots indicating the near and far points that are 1 meter and 3 meters away from the ego vehicle along the lane centerline, respectively.}}
    \label{fig:deepdriving_twopoint}
\end{figure}

Like the baseline setup in Section \ref{baseline}, the near and far points are 1 m and 3 m away from the vehicle along the lane centerline, respectively. The lane centerline can be located by finding the minimum cost at every row in the output image. The near point is therefore the point on the output image that is 15 rows from the bottom of the image, at the minimum cost along that row. The far point is 45 rows from the bottom of the image, at the minimum cost along that row as well. The near and far points are shown in (Fig. \ref{fig:deepdriving_twopoint}). To calculate $\theta_{\rm{near}}$ and $\theta_{\rm{far}}$, we use the following approximations:
\begin{equation}
\theta_{\star}\approx 
\arctan{\left(\frac{W_{td}/2 - col_{\star}}{row_{\star}}\right)},
\quad \star = \mathrm{far,near}
\end{equation}
where $W_{td}$ is the width of the top-down lane prediction output image, $col_{\rm near}$ and $col_{\rm far}$ are the columns of the near and far points respectively, and $row_{\rm near}$ and $row_{\rm far}$ are the rows of the near and far points respectively.

\subsubsection{Vehicle Detection} \label{vehicle_detection_setup}

\begin{figure}[th]
	\centering 
	\includegraphics[width=0.6\columnwidth]{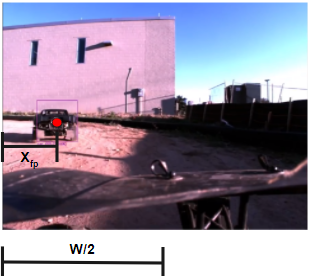}
    \caption{\small{Calculation of $\theta_{\rm{far}}$ using camera FOV and pixel coordinate of vehicle in view.}}
    \label{fig:theta_Far}
\end{figure}

Vehicle detection can be used to identify $\theta_{\rm{far}}$.
The angle $\theta_{\rm{far}}$ is defined as the angle between the vehicle's heading and a faraway point of interest.
In the implementation of the two-point visual controller with YOLOv2, the point of interest is the center of the leading car in the field of view of the car being controlled.
From the visual vehicle detection process described in Section \ref{vehicle_detection}, we obtain a bounding box. 
The pixel at the center of this box will be used as the faraway point of interest. 
The angle to this point can be approximated as
\begin{equation}
\theta_{\rm{far}}\approx\frac{x_{fp}-W_{vd}/2}{\textrm{FOV}/2}, 
\end{equation}
where $x_{fp}$ is the x-coordinate of the faraway point pixel, $W_{vd}$ is the width of the vehicle detection image, and FOV is the field of view of the camera (Fig.~\ref{fig:theta_Far}). 

\section{Experiments and Results}
This section provides the experiments and results of controlling the AutoRally platform using the TPVDCM.

We first tested the line-following performance of the proposed approach. 
In the first experiment, $\theta_{\rm{near}}$ and $\theta_{\rm{far}}$ are calculated as described in Section \ref{baseline}.
In the second experiment, $\theta_{\rm{near}}$ and $\theta_{\rm{far}}$ are calculated as described in Section \ref{top_down_setup}.

We then tested the vehicle-following performance in the third experiment, where a leading vehicle driving along the track corresponds to the ``far" view point, while we control the trailing vehicle.
Thus, the computation of $\theta_{\rm{far}}$ is done using the vehicle detection method described in Section \ref{vehicle_detection_setup}, while $\theta_{\rm{near}}$ is calculated using the top-down lane prediction method in Section \ref{top_down_setup}. 

For each experiment, the vehicle drives five laps counter-clockwise around the track with a target velocity of 4~m/s.
During the tests, a PID controller keeps the vehicle driving at a constant speed, while the TPVDCM with the parameters in Table I steers the vehicle. 
These parameters were initially selected to be similar to those identified in \cite{zafeiropoulos2014twopoint}, although the values were further tuned manually while controlling the vehicle on the track in order to achieve smoother performance.
All computation is run onboard the AutoRally platform.

\begin{table}[ht]
\caption{\small{TPVDCM parameters used in the experiments.}}
\centering
\begin{tabular}{|c|c|}
\hline
Parameter & Value\\
\hline
$T_N$ [sec] & 0.12\\
\hline
$T_P$ [sec] & 0.06\\
\hline
$K_a$ & 30.0\\
\hline
$K_c$ & 10.0\\
\hline
$T_L$ [sec] & 2.8\\
\hline
$T_I$ [sec] & 0.18\\
\hline
\end{tabular}
\vspace{-4mm}
\end{table}

\subsection{Centerline Following Task: Baseline}

\begin{figure}[th]
\centering
\begin{subfigure}{.5\columnwidth}
\centering
\includegraphics[width=\columnwidth]{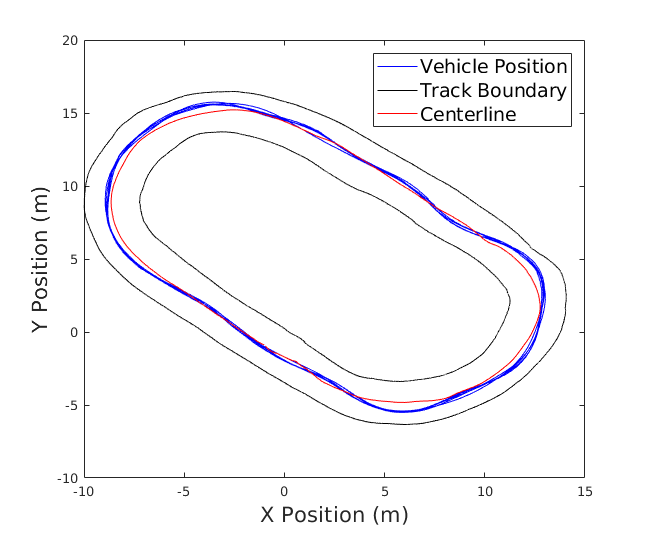}
\end{subfigure}%
\begin{subfigure}{.5\columnwidth}
\centering
\includegraphics[width=\columnwidth]{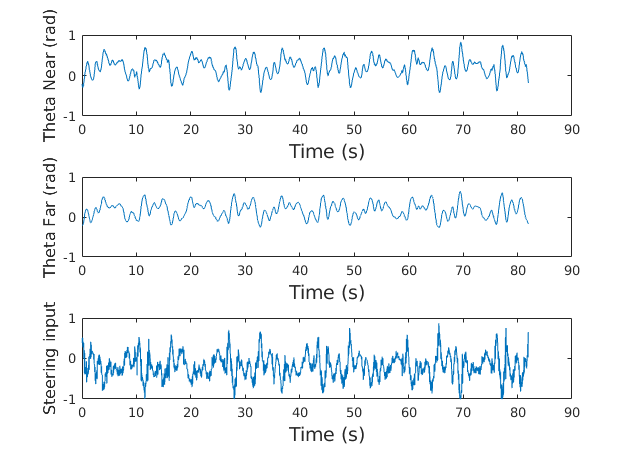}
\end{subfigure}\\[1ex]
(a) \hspace{100pt} (b)
\caption{\small{(a) Vehicle trajectory during baseline centerline following task without vision, driving counter-clockwise. (b) Measured feature-input values ($\theta_{\rm{near}}$, $\theta_{\rm{far}}$) and steering command.}}
\label{fig:baseline_centerline}
\end{figure}

The first experiment verifies the line-following performance of the controller provided with ground truth measurements of feature-input values.
As illustrated in Fig.~\ref{fig:baseline_centerline}(a), the vehicle successfully followed the centerline of the track without colliding with the track boundary. 
The mean absolute lateral error of the vehicle position from the lane centerline during this test was 0.2429 meters, with a standard deviation of 0.1532 meters. 
Fig. \ref{fig:baseline_centerline}(b) depicts the controller's inputs ($\theta_{\rm{near}}$ and $\theta_{\rm{far}}$) and output (steering command) during the experiment.

\subsection{Centerline Following Task: Visual}

\begin{figure}[th]
\centering
\begin{subfigure}{.5\columnwidth}
\centering
\includegraphics[width=\columnwidth]{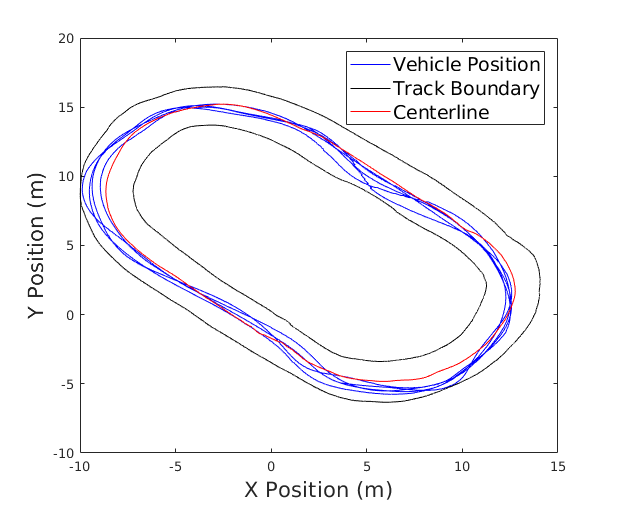}
\end{subfigure}%
\begin{subfigure}{.5\columnwidth}
\centering
\includegraphics[width=\columnwidth]{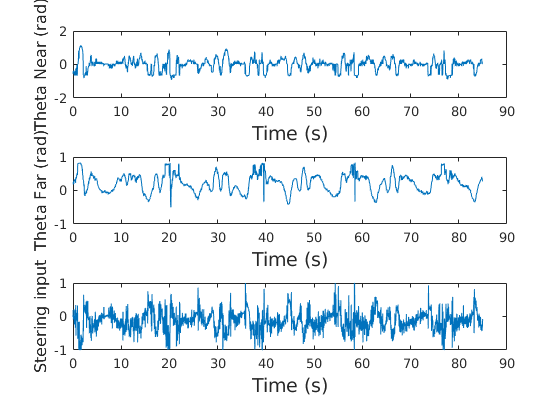}
\end{subfigure}\\[1ex]
(a) \hspace{100pt} (b)
\caption{\small{(a) Vehicle trajectory during centerline following task using the top-down lane cost map prediction network to estimate $\theta_{\rm{near}}$ and $\theta_{\rm{far}}$, driving counter-clockwise. (b) Measured feature-input values ($\theta_{\rm{near}}$, $\theta_{\rm{far}}$) and steering command.}}
\label{fig:vision_centerline}
\end{figure}

In the second experiment, the vehicle extracted feature-input values using the top-down lane cost map prediction network and successfully followed the centerline of the track without colliding with the track boundary, shown in Fig.~\ref{fig:vision_centerline}(a).
The mean absolute lateral error of the vehicle position from the lane centerline during this test was 0.3905 m, with a standard deviation of 0.2840 m. 
Fig. \ref{fig:vision_centerline}(b) depicts the controller's inputs ($\theta_{\rm{near}}$ and $\theta_{\rm{far}}$) and output (steering command) during the experiment.

\subsection{Vehicle Following Task}

\begin{figure}[th]
\centering
\begin{subfigure}{.5\columnwidth}
\centering
\includegraphics[width=\columnwidth]{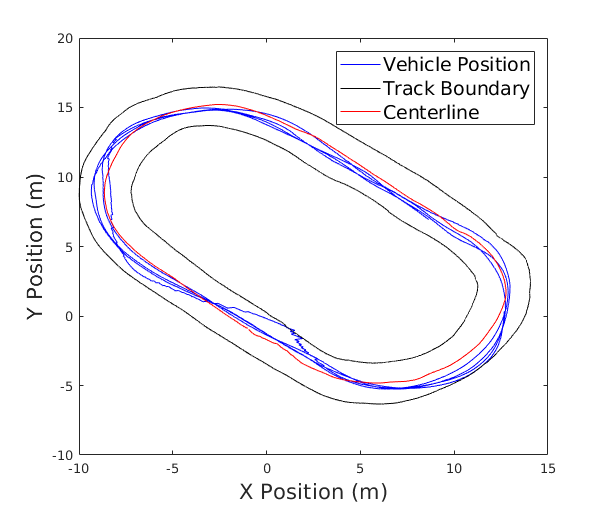}
\end{subfigure}%
\begin{subfigure}{.5\columnwidth}
\centering
\includegraphics[width=\columnwidth]{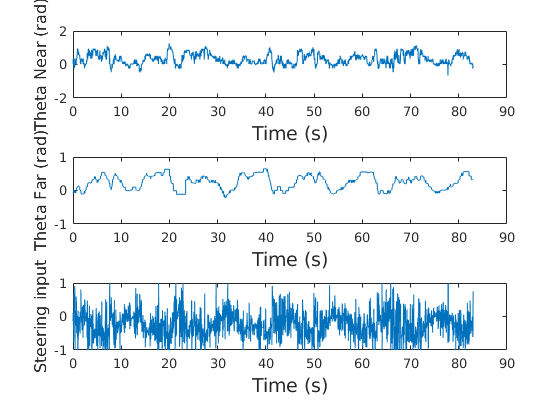}
\end{subfigure}\\[1ex]
(a) \hspace{100pt} (b)
\caption{\small{(a) Trailing vehicle trajectory during vehicle-following task using the top-down lane cost map prediction network to estimate $\theta_{\rm{near}}$ and vehicle detection to estimate $\theta_{\rm{far}}$, driving counter-clockwise. (b) Measured feature-input values ($\theta_{\rm{near}}$, $\theta_{\rm{far}}$) and steering command.}}
\label{fig:vision_detect}
\end{figure}

In this experiment, we evaluated the vehicle-following performance of the controller. 
The leading vehicle was driven manually by a human driver at approximately 4 m/s, while a trailing vehicle drove autonomously using the proposed approach. 
The trailing vehicle used the approach discussed in Section~IV-B to detect the leading vehicle and estimate $\theta_{\rm{far}}$ as discussed in Section \ref{vehicle_detection_setup}.
Note that the leading vehicle does not necessarily drive on the centerline of the track because the centerline is not drawn on the track surface and is thus not visible to the human driver.
As illustrated in Fig.~\ref{fig:vision_detect}(a), the trailing vehicle successfully follows the leading vehicle without colliding with the track boundary.
In this experiment, the mean absolute lateral error from the lane center was 0.4048 m, with a standard deviation of 0.2475 m.
Also, Fig.~\ref{fig:vision_detect}(b) depicts the controller's inputs ($\theta_{\rm{near}}$ and $\theta_{\rm{far}}$) and output (steering command).

\section{Summary}

In this paper, we experimentally demonstrated that using deep-learning-based image processing, a human driver control model can extract feature-input values from driver-point-of-view images and steer a real vehicle.
Specifically, we used the TPVDCM with a top-down lane cost map prediction network and the YOLOv2 network on a 1/5th-scale autonomous vehicle platform.
The TPVDCM is derived from behavioral studies of human drivers. 
Our experiments indicate that the TPDVCM can be used to design a controller mimicking human behavior to control a vehicle autonomously using only visual inputs. 
Our experiments also show that different feature-input points can be used for the TPVDCM, and that a vehicle can be driven using visual inputs to the TPVDCM while maintaining lateral stability.

Since the TPVDCM models human driving behavior during ``normal" driving regimes, in this work we limit the speed of the vehicle to 4 m/s.
For future work with human-inspired vehicle controllers, it may be promising to use a hybrid model \cite{zhang2015computationally} or a model with more sophisticated anticipatory control channel \cite{okamoto2016} to control the vehicle at higher speeds.

\bibliographystyle{IEEEtran}
\bibliography{ref}

\end{document}